\documentclass[letterpaper, 10 pt, conference]{ieeeconf}  % Comment this line out if you need a4paper

\IEEEoverridecommandlockouts                              % This command is only needed if 
\usepackage{hyperref}
\usepackage{url}
\usepackage{graphics} % for pdf, bitmapped graphics files
\usepackage{times} % assumes new font selection scheme installed
\usepackage{booktabs}       % professional-quality tables
\usepackage{amsfonts}       % blackboard math symbols
\usepackage{nicefrac}       % compact symbols for 1/2, etc.
\usepackage{microtype}      % microtypography
\usepackage{algorithm,multirow,xcolor}
\usepackage{algorithmicx}
\usepackage{algpseudocode}
\usepackage{mathtools}
\usepackage{graphicx}
\usepackage{cases}
\usepackage{array}
\usepackage{color}
\usepackage{float}
\usepackage{amssymb}

\usepackage{wrapfig}
\usepackage{subfigure}
\usepackage{amsmath}
\usepackage{amsthm, amssymb}
\theoremstyle{definition}

\usepackage{soul}

\usepackage{graphicx} 
\usepackage{subfigure}
\usepackage{epstopdf}
\title{\LARGE \bf
	Reinforcement Learning Compensated Extended Kalman Filter\\ for Attitude Estimation
}

\author{Yujie Tang$^{1}$, Liang Hu$^{2}$, Qingrui Zhang$^{3}$ and Wei Pan$^{1}$% <-this % stops a space
\thanks{$^{1}$Y.~Tang and W.~Pan are with  the Department of Cognitive Robotics, Delft University of Technology, Netherlands. For correspondence: \texttt{wei.pan@tudelft.nl}.}
\thanks{$^{2}$L.~Hu is with the School of Computer Science and Electronic Engineering, University of Essex, UK.}%
\thanks{$^{3}$Q.~Zhang is with the School of Aeronautics and Astronautics, Sun Yat-Sen University, China. }
}

\begin{document}
	
	\maketitle
	\thispagestyle{empty}
	\pagestyle{empty}

	%%%%%%%%%%%%%%%%%%%%%%%%%%%%%%%%%%%%%%%%%%%%%%%%%%%%%%%%%%%%%%%%%%%%%%%%%%%%%%%%
	\begin{abstract}
		
Inertial measurement units are widely used in different fields to estimate the attitude. Many algorithms have been proposed to improve estimation performance. However, most of them still suffer from 1) inaccurate initial estimation, 2) inaccurate initial filter gain, and 3) non-Gaussian process and/or measurement noise. This paper will leverage reinforcement learning to compensate for the classical extended Kalman filter estimation, i.e., to learn the filter gain from the sensor measurements. We also analyse the convergence of the estimate error. The effectiveness of the proposed algorithm is validated on both simulated data and real data.
	\end{abstract}

	%%%%%%%%%%%%%%%%%%%%%%%%%%%%%%%%%%%%%%%%%%%%%%%%%%%%%%%%%%%%%%%%%%%%%%%%%%%%%%%%
	\section{INTRODUCTION}
ss
	Inertial measurement units (IMU) are widely used to provide accurate attitude estimation in many fields, including: aerospace \cite{8593616}, robotics \cite{8206082} and human motion analysis \cite{h8206016}. Even though the gyroscope alone can compute the sensor's orientation by integrating the angular velocity over time, it suffers from accumulated errors, especially in the drift of angular estimation \cite{sabatelli2011sensor}. A data fusion approach using additional sensors such as an accelerometer and magnetometer is often adopted to achieve higher attitude estimation accuracy. A variety of estimation methods such as the extended Kalman filter (EKF) \cite{vitali2020robust}, unscented Kalman filter (UKF) \cite{7271681} and complementary filter (CF) \cite{5899185} have been proposed, which compute a more reliable estimate using the data collected from all available sensors.  
	
	The EKF and UKF, as two variants of the Kalman filter, use linearisation and deterministic samplings methods, respectively, to obtain a more accurate estimate for nonlinear dynamic systems. The CF uses an optimised gradient descent algorithm to compute the gyroscope measurement error direction as a quaternion derivative \cite{madgwick2011estimation}. Nonetheless, all three methods are sensitive to the initial state estimate. An accurate initial state estimate can lead to the estimation's fast convergence, but the initial estimate is typically chosen empirically, and thus, it is hard to guarantee its accuracy. Furthermore, a filter gain tuning procedure is usually required when deploying filter algorithms in real-world systems. Due to approximation or numerical optimisation methods used in calculating the filter gain, the optimality of the estimation algorithms does not hold, necessitating manual tuning of the filter gains. Moreover, the motion acceleration of the sensor may align with the gravity direction sporadically. This will change the level of measurement noise at some time instants/intervals. Using the classical estimation methods without tuning the filter gain will suffer from slow convergence or even divergence in estimation performance.
	\begin{figure}
		\centering
		\includegraphics[scale=0.3]{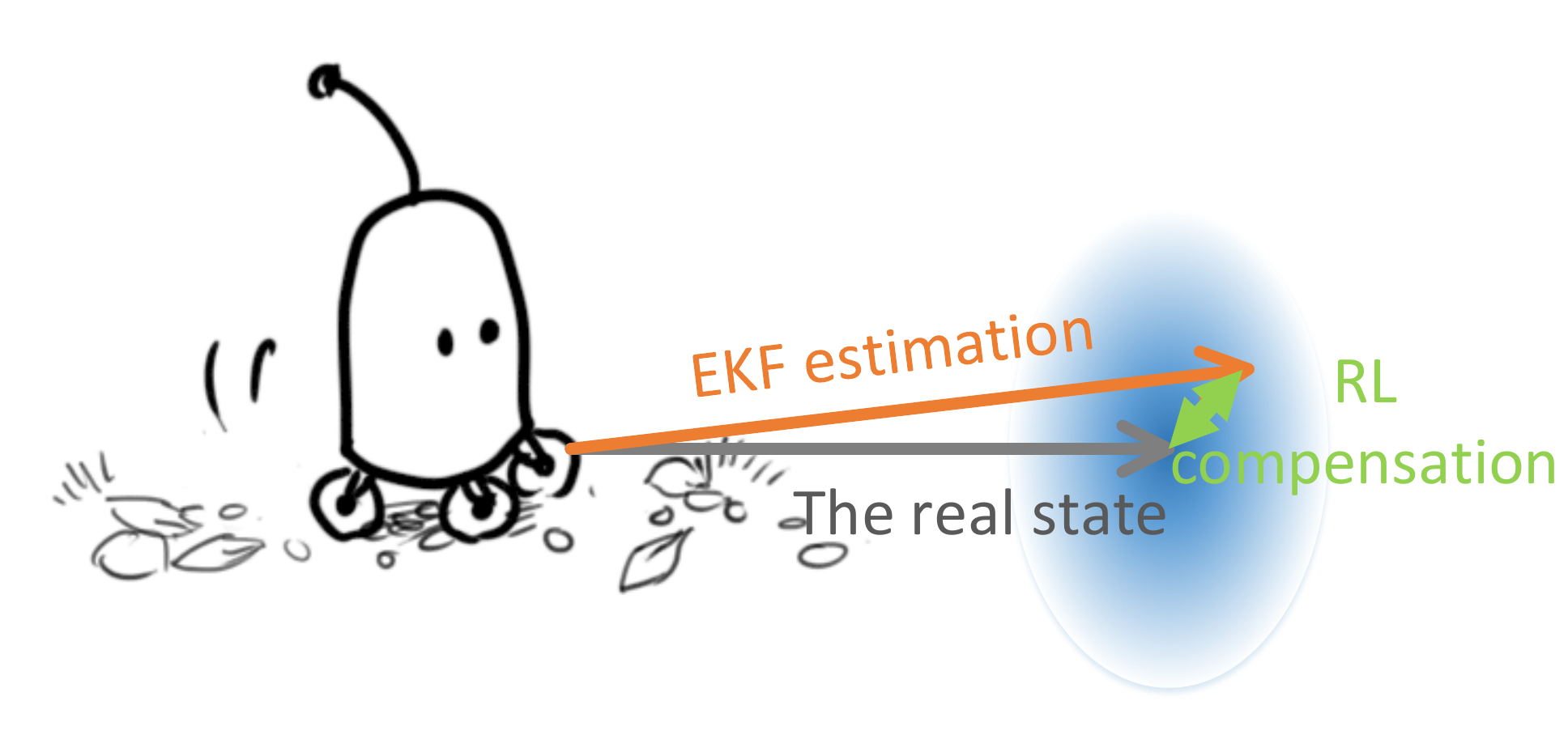}
\vspace{-0.5cm}
		\caption{A cartoon illustration of the extended Kalman filter with reinforcement learning compensation. }
		\label{fig:animation}
		\vspace{-0.5cm}
	\end{figure}
	Several approaches have been proposed to tune the filter gain. \cite{jin2014adaptive}
	proposed a fuzzy processing method to improve the convergence rate. \cite{hyde2008estimation} proposes a switching architecture to prioritise the measurement of sensors in different working conditions to yield robust performance. Learning-based methods are also used to calibrate the IMU parameters and compute gyro corrections that filter undesirable errors in the raw IMU signals \cite{brossard2020denoising}.
	
	This paper tries to combine reinforcement learning (RL) with the EKF to design a tuning-free estimation algorithm insensitive to inaccurate initial state estimate and filter gain. It leverages the merits of both data-driven RL and model-based probabilistic methods instead of only using RL to train the filter gain \cite{hu2020lyapunov, tang2021icra}. The key idea is to add a learnable policy using RL on top of a referenced gain using EKF. Specifically, an RL policy is trained to compute the gyro correction according to the estimation residuals. Then the learned RL policy is applied and acts as a supplement correction based on the EKF. Intuitively, the learned RL policy compensates strongly when the EKF performs bad (big estimate residual) and remains ``idle" if the EKF performs well. Thanks to RL, there is no need to tune the filter parameters in our proposed method manually. As an additional benefit of the added RL in the estimation, the proposed method even works well with non-Gaussian noise, as shown in the experiments. On the other hand, compared with \cite{hu2020lyapunov, tang2021icra} where a pure RL based algorithm is trained/learned from scratch, our proposed method uses the EKF to provide a good starting point of RL training and hence is potentially more sample efficient. A similar idea has recently been explored for control applications \cite{zhang2020model,hwangbo2017control}.

	To summarise, this work aims to design a tuning-free attitude estimation algorithm that maintains good performance even under inaccurate initial state estimates and/or filter gains. We have proposed a double-stage fusion architecture to correct the gyro drift leveraging both EKF and RL methods (as shown in Fig.~\ref{fig:animation}). The proposed estimation method shows superior performance compared with the usual methods in all three scenarios: inaccurate initial state estimate, inaccurate filter gain, and even non-Gaussian noises. Experiments with real data further validate the effectiveness of the proposed method. The rest of the paper is organised as follows. In section \ref{s2}, we describe how the general EKF works in the attitude estimation. Section~\ref{s3} describes the additional RL-based correction step and presents the entire pipeline of the proposed filter. We also analyse the convergence of the estimate error. Experiments and discussion are given in Section~\ref{s4}. 
	
	\section{Extended Kalman filter for Sensor fusion\label{s2}}
	In this paper, the inertial sensors (3D gyroscopes and 3D accelerometers) combined with the magnetometer are used to estimate the attitude. The gyroscopes calculate the attitude movements by integrating the measured angular velocities. The accelerometers and magnetometers observe the local magnetic and gravity direction to estimate the sensor frame's attitude relative to the earth frame. Due to the accumulated error in the integration process and the measurement noise, a Kalman filter is widely used to fuse the separate sensor data. More details can be found in \cite{kok2017using}. 	
	
	\subsection{Orientation from angular velocity}
	
	A tri-axis gyroscope measures the angular velocity along the $x$, $y$, $z$ axes of the sensor frame, termed $y_{\omega} = [ y_{\omega x} \enspace y_{\omega y} \enspace y_{\omega z} ]$. The dynamics of the quaternion are given as 
\vspace{-0.2cm}
	\begin{equation}\label{eq:quatDeviation}
		\dot{q}^\text{nb}_t = \tfrac{1}{2} q^\text{nb}_{t-1} \otimes y_{\omega,t} ,
	\end{equation} 
	where \text{n} and \text{b} stand for the \underline{n}avigation frame and the \underline{b}ody frame respectively, the $\otimes$ operate denotes a quaternion product, $y_{\omega,t}$ is the angular velocity measured at time $t$.
	Accumulating rotation overtime is typically done by discretisation, i.e.,
	\begin{equation}\label{eq:quatIntegration}    
		\begin{split}
			q^\text{nb}_{t} 
			& = q^\text{nb}_{t-1} + T \cdot \dot{q}^\text{nb}_t \\
			& = q^\text{nb}_{t-1} + \tfrac{T}{2} q^\text{nb}_{t-1} \otimes y_{\omega,t} \\
			& = q^\text{nb}_{t-1} \otimes \exp \left( \tfrac{T}{2} y_{\omega,t} \right)
		\end{split}
	\end{equation}
	where $T$ denotes the sampling period, $\exp(\cdot)$ corresponds to the exponential function of the quaternion. This ``integration'' procedure is known to be very sensitive to the measurement noise of the angular velocities.

	\subsection{Orientation from vector observations}
	In attitude estimation, it is typically assumed that the accelerometer only measures the gravity and a magnetometer only measures the earth's magnetic field \cite{madgwick2011estimation}. With the direction of an earth's field known in the earth frame $y^\text{n}$, a measurement of the field's direction within the sensor frame $y^\text{b}$ will allow an orientation of the sensor frame relative to the earth frame $q^\text{nb}$ to be calculated. A determined direction of the earth's field can be expressed in the sensor's frame:
	\begin{equation}\label{eq:OriFromVec}
		\vspace{-0.2cm}
		y^\text{b}
		= (q^\text{nb})^{\star} \otimes y^\text{n} \otimes q^\text{nb}
		= R^\text{nb} \cdot y^\text{n}
	\end{equation}
	where $^{\star}$ denotes conjugate of the quaternion, $y^\text{n}$ is a direction vector in the earth frame that represents either the direction of gravity or that of the magnetic field. The corresponding description of $y^\text{n}$ in the sensor frame is denoted as $y^\text{b}$. And $R^\text{nb}$ is the rotation matrix associated with the orientation $q^\text{nb}$. 
	
	For any single measurement $y^\text{b}$, there will not be a unique orientation solution to the under-determined function $h(y^\text{b}, y^\text{n},q^\text{nb})=0$ (according to Eq.~\eqref{eq:OriFromVec}). Thus, the gravity and magnetic observations are used to reference the Tilt-Pitch angle and the yaw angle, respectively. While both the accelerometer and magnetometer measurements are often contaminated by large measurement noise under some working conditions \cite{madgwick2011estimation}, they need to be fused with another stable measurement source, e.g., a gyroscope, to achieve better performance. 
	
	\subsection{Extended Kalman filter for attitude estimation}
	The EKF is widely used in attitude estimation to fuse the measurements from gyroscopes, accelerometers, and magnetometers for a single, accurate estimate of the orientation. The orientation is estimated recursively by performing a \textit{prediction update} and a \textit{correction update}. The \textit{prediction update} use model \eqref{eq:quatIntegration} to predict the state (orientation estimation in terms of quaternion) of the next time step as follows
	\begin{equation}
		\begin{aligned}
			\label{eq:ori_qEkf_timeUpdate}
			\tilde{q}^\text{nb}_{t \mid t-1} &= \hat{q}^\text{nb}_{t-1 \mid t-1} \otimes \exp \left( \tfrac{T}{2} y_{\omega,t-1} \right), \\
			P_{t \mid t-1} &= F_{t-1} P_{t-1 \mid t-1} F_{t-1}^\top + G_{t-1} Q G_{t-1}^\top,
		\end{aligned}
	\end{equation}
	with $Q = \Sigma_\omega$,
	$F_{t-1} = \left( \exp (\tfrac{T}{2} y_{\omega,t-1}) \right)^{R} $ and $G_{t-1} = -\tfrac{T}{2} \left( \hat{q}^{\text{nb}}_{t-1 \mid t-1} \right)^{L} \tfrac{\partial \exp (e_{\omega,t-1})}{\partial e_{\omega,t-1}}$.
	$\Sigma_\omega$ is the covariance matrix of the gyroscope measurement noise, $(\cdot)^L$ and $(\cdot)^R$ are the left- and right- quaternion-product matrices respectively \cite{sola2017quaternion}.
	
	By using \eqref{eq:OriFromVec}, the measurements from the accelerometer and magnetometer are fused to correct the  state predictions. The \textit{correction update} equations of the EKF are as follows:
	\begin{equation}
		\begin{aligned}
			\tilde{q}^\text{nb}_{t \mid t} &= \tilde{q}^\text{nb}_{t \mid t-1} + K_t \varepsilon_t,  \\
			\tilde{P}_{t \mid t} &= P_{t \mid t-1} - K_t S_t K_t^\top, 
			\label{eq:ori_qEkf_measUpdate1}
		\end{aligned}
	\end{equation}
	\hspace{-0.01cm}where
	{ $\varepsilon_{t} \triangleq y_{t} - y_{t \mid t-1}$, $S_{t} \triangleq H_{t} P_{t \mid t-1} H_{t}^\top + R$, $K_{t} \triangleq P_{t \mid t-1} H_{t}^\top S_{t}^{-1}$,
		$
		y_t = \begin{pmatrix} y_{\text{a},t} \\ y_{\text{m},t} \end{pmatrix}$, 
		$ 
		y_{t \mid t-1} = \begin{pmatrix} -\tilde{R}^\text{bn}_{t \mid t-1} g^\text{n} \\ \tilde{R}^\text{bn}_{t \mid t-1} m^\text{n} \end{pmatrix}$, 
		$H_t = \begin{pmatrix} - \left. \tfrac{\partial \tilde{R}^\text{bn}_{t \mid t-1}}{\partial \tilde{q}^\text{nb}_{t \mid t-1}} \right|_{\tilde{q}^\text{nb}_{t \mid t-1}=\tilde{q}^\text{nb}_{t \mid t-1}} g^\text{n} \\ \left. \tfrac{\partial \tilde{R}^\text{bn}_{t \mid t-1}}{\partial \tilde{q}^\text{nb}_{t \mid t-1}} \right|_{\tilde{q}^\text{nb}_{t \mid t-1}=\tilde{q}^\text{nb}_{t \mid t-1}} m^\text{n} \end{pmatrix}$,  
		$R = \begin{pmatrix} \Sigma_\text{a} & 0 \\ 0 & \Sigma_\text{m} \end{pmatrix}$}.
	$\Sigma_\text{a}$ and $\Sigma_\text{m}$ denote the covariance matrix of the measurement noise of accelerometer and magnetometor respectively,  $g^\text{n}$ and $m^\text{n}$ are the accelerometer and magnetometor measurements at time $t$ respectively.

	However, there are a few drawbacks of the EKF and its variants.
	First, the linearisation error is inevitable \cite{8009830}, since the EKF is based on the first-order approximation of the nonlinear models (Eq.~\eqref{eq:quatDeviation}). This means an inaccurate initial estimate $q^\text{nb}_{0 \mid 0}$, i.e., the point of the first linearisation is undesirable. 
	Second, the filter gain computed using inaccurate noise covariance in the EKF, needs to be tuned a priori.
	Third, the EKF is derived based on the assumption that the measurement noise distribution is Gaussian and time-invariant, which are often too limited in practice. For example, the covariance of measurement noise may change abruptly over time.
	To address these issues, we propose a reinforcement learning (RL) approach to compensate for the estimation obtained from the EKF.
	\begin{figure}
		\centering
		\includegraphics[scale=0.6]{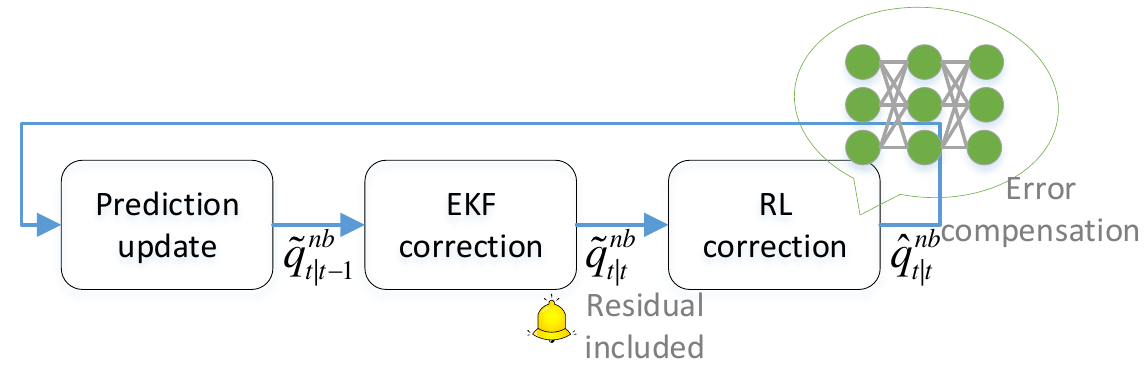}
		\vspace{-0.3cm}
		\caption{RL compensated EKF workflow. First, gyroscope measurements are used to predict the attitude using \eqref{eq:ori_qEkf_timeUpdate}. Then an EKF correction is performed to update the prediction using \eqref{eq:ori_qEkf_measUpdate1}. Last, a RL policy is further used to compensate the residuals using \eqref{eq:rlupdate_q}.}
		\label{fig:TwoCorrection}
		\vspace{-0.5cm}
	\end{figure}

	\section{RL Compensated EKF Algorithm}
	\label{s3}

	\subsection{Preliminary to Reinforcement Learning }
	Reinforcement learning (RL) is a class of learning algorithms in which an agent interacts with an environment to find an action selection policy and optimise its long-term performance \cite{sutton2018reinforcement}. The interaction is conventionally modelled as a Markov Decision Process (MDP) \cite{puterman1990markov}. 
	
	A MDP can be denoted as a tuple $<\mathcal{S},\mathcal{A}, \mathcal{P}, \mathcal{R},\gamma >$, where $\mathcal{S}$ is the state, $\mathcal{A}$ is the action sampled from a stochastic policy $\pi$, $\mathcal{P}:\mathcal{S} \times \mathcal{A} \times \mathcal{S}\to \mathbb{R}$ defines a transition probability, $\mathcal{R}: \mathcal{S} \times \mathcal{A} \to \mathbb{R}$ is a reward function, and $\gamma \in [0,1]$ is a discount factor. RL is used to find a policy $\pi (\mathcal{A}_t | \mathcal{S}_t)$ and state-update function $\mathcal{V}_\pi(\mathcal{S}_{t}) = \sum\limits_{t}^\infty \sum\limits_{A_t} \pi (A_t|\mathcal{S}_{t}) \sum\limits_{\mathcal{S}_{t+1}} \mathcal{P}_{t+1|t} (\mathcal{R}_{t}+\gamma \mathcal{V}_{\pi}(\mathcal{S}_{t+1}))$ to maximise the expected sum of discounted rewards. And the policy is the probability of choosing an action $a_t \in \mathcal{A}$ at state $s_t \in \mathcal{S}$ \cite{sutton2018reinforcement}.

	\subsection{RL Problem Formulation}
	
	We first rewrite the residual of the EKF, i.e., $\hat{\eta}^\text{nb} \in \mathbb{R}^3$, on rotation group $SO(3)$ instead of quaternion to drop the unit determinant condition, where $\hat{\eta}^\text{nb} = \tilde{q}^\text{nb} \otimes (\hat{q}^\text{nb})^{\star}$. When it is changed back to quaternions, the exponential map will be used \cite{sola2017quaternion}: 
	\begin{align}
		exp:\; \mathbb{R}^3 \to SO(3);\;\;q_{\hat{\eta}^\text{nb} } = \exp \left( \tfrac{\hat{\eta}^\text{nb} }{2} \right).
	\end{align}
	The high-level plan can be illustrated in Fig.~\ref{fig:TwoCorrection}. 
	In the RL correction module, the difference between the measurements $y_{t}$ from sensors and the estimated observation 
	$ \tilde{y}_{t \mid t} = h(\tilde{q}_{t\mid t}^{\text{nb}}) = \begin{pmatrix} -\tilde{R}^\text{bn}_{t \mid t} g^\text{n} \\ \tilde{R}^\text{bn}_{t \mid t} m^\text{n} \end{pmatrix}$ from the EKF 
	where $h(\cdot)$ is a compact form of \eqref{eq:OriFromVec},
	$\varepsilon_{t}^\text{RL} = y_{t} - \tilde{y}_{t \mid t}$
	is used to obtain the estimate residual $\hat{\eta}_{t}^\text{nb,RL}$ by computing a gain $U_{t}$, 
	in order to get the improved estimate $\hat{q}^\text{nb}_{t \mid t}$. 
	The estimate residual $\hat{\eta}_{t}^\text{nb,RL}$ computed using RL algorithms plays a role of compensating the estimate $\tilde{q}^\text{nb}_{t \mid t}$ obtained from the EKF.
	
	\begin{figure}
		\centering
		\includegraphics[scale=0.45]{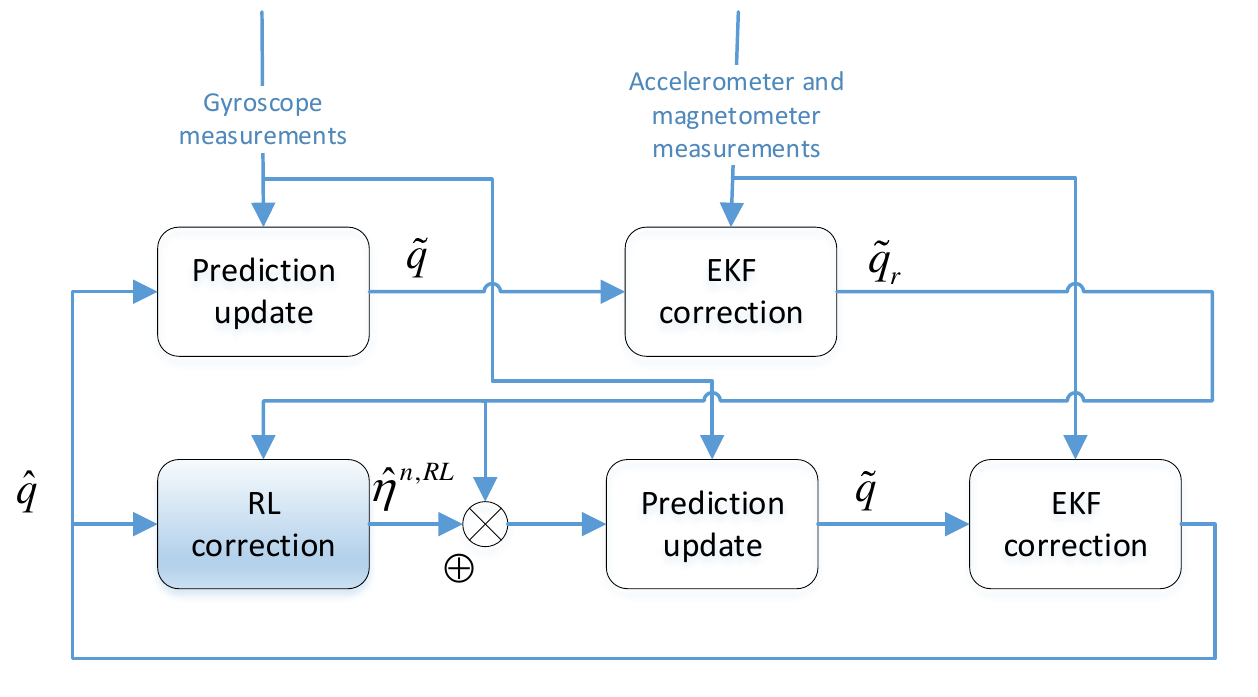}
		%\vspace{-1cm}
		\caption{Schematic illustration of RLC-EKF algorithm.}	\label{fig:filterFlowChart}
		\vspace{-0.5cm}
	\end{figure}
	Then we have the following equations: 
	\begin{equation}
		\begin{aligned}
			\hat{\eta}_{t}^\text{nb,RL} &= U_{t} \varepsilon_{t}^\text{RL}\\
			\hat{q}^\text{nb}_{t \mid t} &= \exp \left( \tfrac{\hat{\eta}_{t}^\text{nb,RL}}{2} \right) \otimes \tilde{q}^\text{nb}_{t \mid t}
			\label{eq:rlupdate_q}
		\end{aligned}
	\end{equation}

	Based on the filter process in~\eqref{eq:rlupdate_q}, we have 
	\vspace{-0.2cm}
	\begin{equation}
		\begin{aligned}
			\hat{\eta}_{t}^\text{nb,RL} &= 2\log \left( \hat{q}^\text{nb}_{t \mid t} \otimes (\tilde{q}^\text{nb}_{t \mid t})^{\star} \right) \\
			&= 2\log \left( ( \exp ( \tfrac{U_{t} \varepsilon_{t}^\text{RL}}{2} ) \otimes \tilde{q}^\text{nb}_{t \mid t} ) \otimes (\tilde{q}^\text{nb}_{t \mid t})^{\star} \right)
		\end{aligned}
	\end{equation} 
	where $\tilde{q}^\text{nb}_{t \mid t} = f (  \hat{q}^\text{nb}_{t-1 \mid t-1} ) = \exp \left( \tfrac{\hat{\eta}_{t-1}^\text{nb,RL}}{2} \right) \otimes \tilde{q}^\text{nb}_{t-1 \mid t-1}$,
	and $f(\cdot)$ denotes the estimation correction step in the EKF.

	The dynamics of the residual $\hat{\eta}_{t}^\text{nb,RL}$ can be characterised  as an MDP, in which the stochasticity essentially comes from the sensor noise: 
	\begin{equation}\label{eq:MDP}
		\hat{\eta}_{t}^\text{nb,RL} \sim \mathcal{P}\left(\hat{\eta}_{t}^\text{nb,RL} | \hat{\eta}_{t-1}^\text{nb,RL}, U_{t}\right), \forall t \in \mathbb{Z}_{+},
	\end{equation}
	where the estimate residual $\hat{\eta}_{t}^\text{nb,RL} \in \mathcal{S}$ is the state, the estimated gain $U_{t} \in \mathcal{A}$ is the action sampled from the trained policy, $\mathcal{P}(\hat{\eta}_{t}^\text{nb,RL}|\hat{\eta}_{t-1}^\text{nb,RL}, U_{t})$ indicated the transition probability function of the estimation.

	To this end, the attitude estimation problem can be formulated as an RL problem, i.e., learn a policy $U_t$ to control the estimate residual $\hat{\eta}_{t}^\text{nb,RL}$.

	\begin{algorithm} 
		\caption{RL Compensated EKF (RLC-EKF) Algorithm for Orientation Estimation } 
		\label{alg:ALG1}
		\begin{flushleft}
			\textbf{INPUTS:} Initial data $\{y_{w,t}, y_{a,t}\}^N_{t=1}$, magnetometer data $\{y_{m,t}\}$, and the initial estimation of the orientation $q^\text{nb}_{0 \mid 0}$, \\
			\textbf{OUTPUTS:} Improved estimation of the orientation $\hat q^\text{nb}_{t|t}$.
		\end{flushleft}
		\begin{algorithmic}[1]
			\For{t=1,2,3...N}
			
			{ \textit{Prediction Update}} 
			
			\State Apply~\eqref{eq:ori_qEkf_timeUpdate} with angular velocity.
			
			{ \textit{EKF Correction}:}
			\State  Correction update using $y_{a,t}$ and $y_{m,t}$ as in~\eqref{eq:ori_qEkf_measUpdate1}
			
			{ \textit{Quaternion Normalisation}:}  
			\State  Normalise the quaternion and its covariance as
			$\tilde{q}_{t \mid t}^\text{nb} = \tfrac{\tilde{q}^\text{nb}_{t \mid t}}{\| \tilde{q}^\text{nb}_{t \mid t}\|_2}$,
			$\tilde P_{t \mid t}^\text{EKF} = J_t \tilde{P}_{t \mid t} J_t^\top$
			with $J_t = \tfrac{1}{\| \tilde{q}^\text{nb}_{t \mid t} \|_2^3} \tilde{q}^\text{nb}_{t \mid t} \left( \tilde{q}^\text{nb}_{t \mid t} \right)^\top$.

			{\textit{RL Correction - estimate the residual in estimation}:} 
			\State Compute the residual $\hat{\eta}_t^\text{nb,RL}$ and gain $U_t$ using Algorithm~\ref{alg:RL}, where $\hat{\eta}_t^\text{nb,RL} = U_t \varepsilon_t^\text{RL},, U_{t} =\pi(\hat{\eta}_{t-1}^\text{nb,RL}), \varepsilon_t^\text{RL} = y_t - y_{t|t}$, $y_t = \begin{pmatrix} y_{\text{a},t} \\ y_{\text{m},t} \end{pmatrix}$, and $y_{t \mid t} = \begin{pmatrix} -\tilde{R}^\text{bn}_{t \mid t} g^\text{n} \\ \tilde{R}^\text{bn}_{t \mid t} m^\text{n} \end{pmatrix}$.

			{\textit{RL Correction -  inject the estimate residual}:} 
			\State Inject the error into the quaternion state and adapt the covariance matrix $\hat{q}^\text{nb}_{t \mid t} = \exp \left( \tfrac{\hat{\eta}_t^\text{nb,RL}}{2} \right) \odot \tilde{q}^\text{nb}_{t \mid t}$ and $\hat{P}_{t \mid t} = M_t \tilde{P}_{t \mid t} M_t^\top$ 
			with $M_t =  \bigg( \exp \left( \tfrac{\hat{\eta}_t^\text{nb,RL}}{2} \right) \bigg) ^{L}$.	
			\EndFor
		\end{algorithmic}
	\end{algorithm}

	\subsection{RL Compensated EKF Algorithm}
	In the RL problem, a cost function\footnote{We use cost function instead of reward function.} $\mathcal{C}(\hat{\eta}_{t-1}^\text{nb,RL}, U_{t}) \in \mathcal{C}$ will be used to measure the goodness of a state-action pair, i.e.,
	\begin{equation}
		\begin{aligned}
			C(\hat{\eta}_{t-1}^\text{nb,RL}, U_{t})=\mathbb{E}_{P(\cdot|\hat{\eta}_{t-1}^\text{nb,RL}, U_{t})}[\|\hat{\eta}_{t}^\text{nb,RL}\|^2]
			\label{eq:costfunction}
		\end{aligned}
	\end{equation}
	and the return is the sum of discounted cost $\sum_{\tau=t}^{\infty}\gamma^{\tau-t}C(\hat{\eta}_{t-1}^\text{nb,RL}, U_{t})$ with the discount factor $\gamma\in [0,~1)$. 
	Our aim is to learn the estimator gain $U_{t} =\pi(\hat{\eta}_{t-1}^\text{nb,RL})$ in \eqref{eq:rlupdate_q} as a policy using RL algorithm. And the policy will be learned/approximated by using a deep neural network. This means any state-of-the-art deep RL algorithms can be readily applied.

	The full pipeline is illustrated in Fig.~\ref{fig:filterFlowChart}. The proposed RL compensated EKF (RLC-EKF) algorithm is summarised in Algorithm~\ref{alg:ALG1}. It should be noted that the superscript $\tilde{\cdot}$ (not $\hat{\cdot}$) indicates the estimate is intermediate and still needs to be updated. More reliable estimation of the orientation can be obtained after the two correction steps, i.e., EKF and RL Correction in Algorithm~\ref{alg:ALG1}.

	\subsection{Convergence of Estimate Error}
	In this attitude estimation task, it is desired that the estimate error $\hat{\eta}_{t-1}^\text{nb,RL}$ converges to a real number as small as possible. The vanilla EKF should be stable to ensure the convergence of the attitude estimation. After introducing the reinforcement learning compensation structure, the final estimate of the proposed filter is expected to be better than the vanilla EKF and should not deteriorate the convergence of EKF. Inspired by the work \cite{zhang2020model1,zhang2020model}, based on EKF, we will show that the mean square of the estimate residual in the RL-compensated filter is guaranteed to converge within a positive bound.

	The value function in reinforcement learning can be written as $\mathcal{V}^i\left(\hat{\eta}_{t}^\text{nb,RL}\right)=-\sum_{\tau=t}^{\infty}\gamma^{\tau-t}C^i(\hat{\eta}_{t-1}^\text{nb,RL}, U_{t}^{i})$. In the presented design, the RL mechanism is introduced to further improve the performance of EKF. Hence, it is reasonable to assume that $\mathcal{V}^i\left(\hat{\eta}_{t}^\text{nb,RL}\right)$ is a continuously differentiable function with
\vspace{-0.07cm}
	\begin{align*}
		\mathcal{V}^i\left(\hat{\eta}^\text{nb,RL}_{t+1}\right)-\mathcal{V}^i\left(\hat{\eta}_{t}^\text{nb,RL}\right)&
		\leq - \mathcal{W}\left(\hat{\eta}_{t}^\text{nb,RL}\right)+\mu^i \label{eq: EKF_Stab1}
	\end{align*}
	where $\mathcal{W}\left(\hat{\eta}_{t}^\text{nb,RL}\right)$ is a continuous positive definite function, $\mu^i>0$, and $i$ denotes the $i$-th iteration of the RL algorithm. There exists $\mathcal{W}\left(\hat{\eta}_{t}^\text{nb,RL}\right) > \mu$. When $i=0$, it corresponds to the vanilla EKF scenario, as the RL policy has negligible initial values around zeros. Next, let's approve it also holds in the later iterations.
	
	Let $U_{t}^i$ be the estimated gain of the $i$-th iteration of RL algorithm. Since $\mathcal{V}^i\left(\hat{\eta}^\text{nb,RL}_{t}\right)=-C^i(\hat{\eta}_{t}^\text{nb,RL}, U_{t}^{i})+\gamma\mathcal{V}^i\left(\hat{\eta}_{t+1}^\text{nb,RL}\right)$, we have
	\begin{align*}
		\left(1-\gamma\right)\mathcal{V}^i\left(\hat{\eta}^\text{nb,RL}_{t+1}\right)& 
		\leq - \mathcal{W}\left(\hat{\eta}_{t}^\text{nb,RL}\right)+\mu^i-C^i(\hat{\eta}_{t}^\text{nb,RL}, U_{t}^{i})
	\end{align*}
	
	In the policy evaluation, the following Bellman backup operation is repeatedly conducted.
	\begin{equation*}
		\mathcal{V}^{i+1}\left(\hat{\eta}^\text{nb,RL}_{t}\right) = -C^{i+1}(\hat{\eta}_{t}^\text{nb,RL}, U_{t}^{i+1})+\gamma \mathcal{V}^{i}\left(\hat{\eta}_{t+1}^\text{nb,RL}\right) \label{eq: iter1_V} 
	\end{equation*}
	In the policy improvement, $U_{t}$ is updated to minimise the discounted accumulated cost, so 
	\begin{align*}
		&\mathcal{V}^{i+1}\left(\hat{\eta}^\text{nb,RL}_{t+1}\right)- \mathcal{V}^{i+1}\left(\hat{\eta}^\text{nb,RL}_{t}\right) \\
		& = C^{i+1}(\hat{\eta}_{t}^\text{nb,RL}, U_{t}^{i+1})+\left(1-\gamma\right)\mathcal{V}^{i+1}\left(\hat{\eta}^\text{nb,RL}_{t+1}\right)\\
		& \leq C^{i+1}(\hat{\eta}_{t}^\text{nb,RL}, U_{t}^{i+1})+\left(1-\gamma\right)\mathcal{V}^i\left(\hat{\eta}^\text{nb,RL}_{t+1}\right)\\
		& \leq  - \mathcal{W}\left(\hat{\eta}_{t}^\text{nb,RL}\right)+\mu^i-C^i(\hat{\eta}_{t}^\text{nb,RL}, U_{t}^{i})+C^{i+1}(\hat{\eta}_{t}^\text{nb,RL}, U_{t}^{i+1})\label{eq: iter2_V}
	\end{align*}
	Hence, the RL algorithm will ensure stable performance for all iterations. At each iteration of the policy improvement, the discounted accumulated cost will be reduced, so $C^{i+1}(\hat{\eta}_{t}^\text{nb,RL}, U_{t}^{i+1})\leq C^i(\hat{\eta}_{t}^\text{nb,RL}, U_{t}^{i})$ and $\mu^i-C^i(\hat{\eta}_{t}^\text{nb,RL}, U_{t}^{i})+C^{i+1}(\hat{\eta}_{t}^\text{nb,RL}, U_{t}^{i+1})\leq \mu^i$. It implies that each iteration will potentially reduce the ultimate bound of $\hat{\eta}_{t}^\text{nb,RL}$. In the implementation, accumulated discounted cost will be approximated by a multiple layer perceptron (MLP) with parameters denoted by $\theta$. The actor network parameters are denoted by $\phi$. 
	The training process of the RL module in Step 5 of Algorithm~\ref{alg:ALG1} is summarised in Algorithm \ref{alg:RL}, in which $J_V$ and $J_\pi$ denote the optimisation objective for the critic and actor objectives \cite{zhang2020model1,zhang2020model}.

	\begin{algorithm}[tbp]
		\caption{RL algorithm for Step 5 in Algorithm~\ref{alg:ALG1}}  \label{alg:RL}
		\begin{algorithmic}[1]
			\State Initialise parameters $\theta$ for the critic $\mathcal{V}_\theta\left(\hat{\eta}_{t}^\text{nb,RL}\right)$, and $\phi$ for the actor network. The initial value of $U_{t}^{0}$ is from the vanila EKF module. Initialise the replay memory $\mathcal{D}\leftarrow\emptyset$.
			\State Assign initial values to the critic parameter $\theta\leftarrow\theta^0$ and its target $\bar{\theta}\leftarrow\theta^0$
			\For{Data collection steps}
			\State Choose an action $U_{t}$ sampled from $ \pi(\hat{\eta}_{t}^\text{nb,RL})$ 
			\State Run the simulator and EKF module \& collect data
			\EndFor
			\For{each gradient update step}
			\State Sample a batch of data $\mathcal{B}$ from $\mathcal{D}$
			\State $\theta\leftarrow\theta-l_V \nabla_{\theta} J_{V}\left(\theta\right) $
			\State $\phi\leftarrow\phi-l_\pi \nabla_{\phi}J_{\boldsymbol{\pi}}\left(\phi\right)$
			\State $\bar{\theta}\leftarrow\kappa\theta+\left(1-\kappa\right)\bar{\theta}$
			\EndFor
		\end{algorithmic}
	\end{algorithm}

	\section{EXPERIMENTAL RESULTS \label{s4}}
	This section will evaluate the proposed algorithm (RLC-EKF) on both simulated and real data. We compare with the EKF~\cite{kok2017using}, the CF~\cite{madgwick2011estimation} and the RLF~\cite{tang2021icra}. We evaluate the performance in three scenarios: (1) inaccurate initial state estimate, (2) inaccurate filter gain, (3) inaccurate noise model. Before going into the details, a quick summary of the feature of the algorithms can be found in Table.~\ref{tab:expSetup}
	
	\begin{table}[htbp]
		\centering
		\caption{
			Features of RLC-EKF compared with other methods.}
		\begin{tabular}{lcccc}
			\hline
			Applicability & EKF & CF & RLF &RLC-EKF \\\hline
			Inaccurate initial estimation  & $\times$ & $\surd$&$\times$&$\surd$\\\hline
			Inaccurate filter gain & $\times$ & $\times$&$\surd$&$\surd$\\\hline
			Inaccurate noise model & $\times$ & $\surd$&$\surd$&$\surd$\\\hline
			
		\end{tabular}
		\label{tab:expSetup}
	\end{table}

	A relatively trivial pattern of the angular velocity (see Fig.2 in our previous work \cite{tang2021icra}) is used as the training profile. The angular velocity profile remains the same in each training episode, while the initial state $q^\text{nb}$ and the initial estimation of the state $\hat{q}^\text{nb}_{0 \mid 0}$ are randomly sampled with a uniform distribution $\mathbb{U}([-1,-1,-1,-1],[1,1,1,1])$. The sampling rate is 100 Hz (consistent with that for real data in Section \ref{sec:realData}). During training, the sensor noise is sampled with the following distribution \cite{kok2017using}: 
	$e_{\omega,t} \sim \mathcal{N}(0,\Sigma_\omega), \Sigma_\omega=0.0003I_{3\times3}, e_{\text{a},t} \sim \mathcal{N}(0, \Sigma_\text{a}), \Sigma_\text{a}=0.0005I_{3\times3}, e_{\text{m},t} \sim \mathcal{N}(0,\Sigma_\text{m}), \Sigma_\text{m}=0.0003I_{3\times3}.
	$
	We independently train 20 policies and select the one with the lowest validate error for inference on unknown profiles.
	The training of the RL policy in RLC-EKF used PPO2~\cite{schulman2017proximal} for Algorithm~\ref{alg:RL} in Section.~\ref{s3}. The hyperparameters are the same as our previous work (see Table.1 in \cite{tang2021icra}).

	\subsection{Results for simulated data}

	\subsubsection{Scenario 1: Inaccurate Initial Estimation}
	
\begin{figure}[h!]
		\centering
		\includegraphics[scale=0.38]{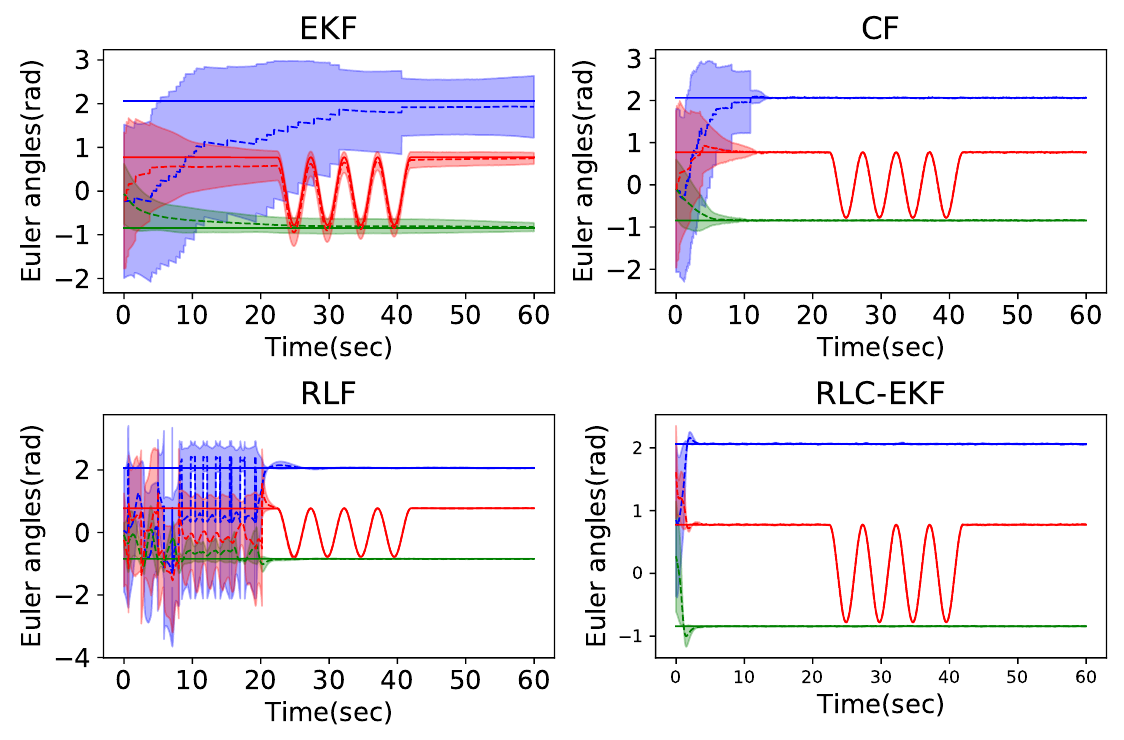}
		\caption{
			The attitude estimation performance of EKF, CF, RLF and RLC-EKF when the initial estimation is randomly selected. The solid and dashed lines respectively represent the ground truth and the average estimate. The shaded areas indicate the standard deviation over 50 independent runs. }
		\label{fig:inferq_onlyRL}
	\end{figure}

	We compare our algorithm with EKF, CF and RLF when the initial estimate $\hat{q}^\text{nb}_{0 \mid 0}$ is inaccurate and randomly sampled with a uniform distribution $\hat{q}^\text{nb}_{0 \mid 0}\sim\mathbb{U}([-1,-1,-1,-1],[1,1,1,1])$. The estimation is expected to converge to the true state as quickly as possible. The hyperparameters for EKF and CF are directly adapted from~\cite{madgwick2011estimation, kok2017using}. The CF gain $\beta$ is set as $0.041$. The adjustable measurement covariance in EKF is chosen as the true sensor covariance. The results of the estimation performance are shown in Fig.~\ref{fig:inferq_onlyRL}. The results indicate that our algorithm can quickly converge to the true state compared with EKF and CF. 
	
	\subsubsection{Scenario 2: Inaccurate Filter Gain}
\begin{figure}[h!]
		\centering
		\includegraphics[scale=0.26]{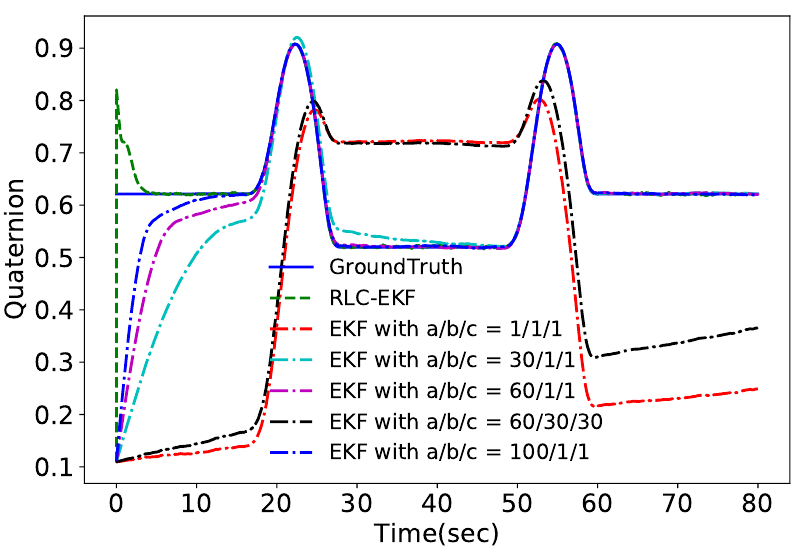}
		~
		\includegraphics[scale=0.26]{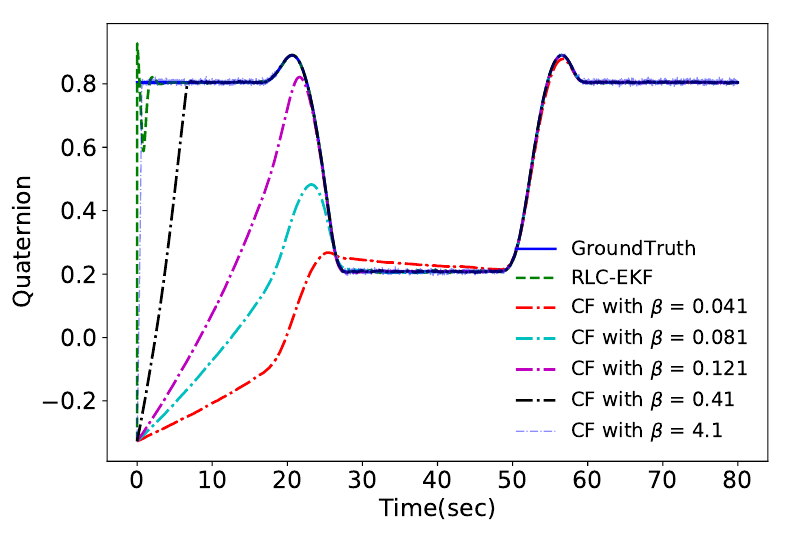}
		\caption{
			The left figure shows the experimental results of EKF with different covariance matrices. We set the estimated measurement covariance of the gyroscope, accelerometer, and magnetometer as their real covariance, respectively times a, b and c. The right figure shows the experimental results of the CF with different initial filter gain, $\beta$. We set $\beta$ as 0.041, 0.081, 0.121, 0.41 and 4.1 respectively. }
		\label{fig:inferq_ekfs}
	\end{figure}

	The performance of the filters largely depends on their filter gain. In CF it is the adjustable parameter $\beta$. 
	In the EKF, since $K_{t} \triangleq P_{t \mid t-1} H_{t}^\top (H_{t} P_{t \mid t-1} H_{t}^\top + R)^{-1}$ where the covariance of measurement noise $R$ needs to be specified/tuned in the beginning, the gain $K_t$ is determined by $R$. 
	Experimental results of EKF and CF with different filter gains can be found in Fig.~\ref{fig:inferq_ekfs}. With different filter gains, their estimation performance varies a lot. It significantly stresses the importance of parameter tuning. 

	\subsubsection{Scenario 3: Inaccurate noise model}
\begin{figure}[ht!]
		\centering
\vspace{-0.2cm}
		\includegraphics[scale=0.395]{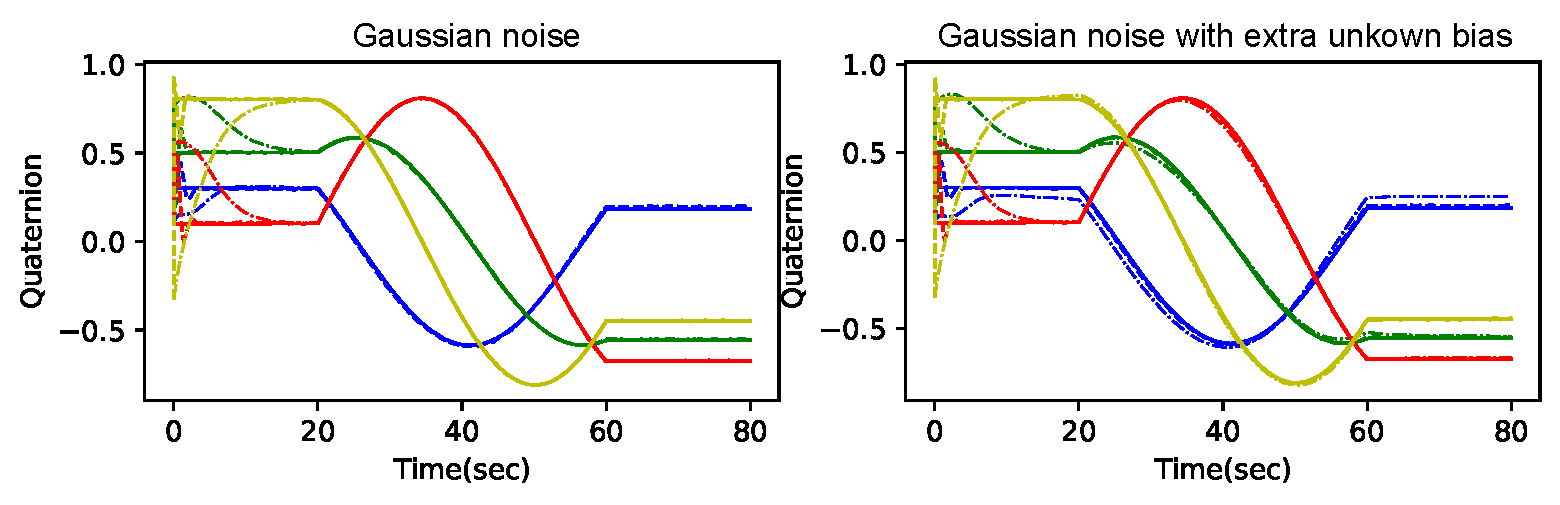}
		\vspace{-0.2cm}
		\caption{
		The attitude estimation performance of RLC-EKF and EKF on measurements with and without an accurate noise model. One only includes a known Gaussian noise. The other includes a known Gaussian noise with unknown extra bias. The solid, dashed, and dash-dot lines represent the ground truth, the RLC-EKF estimation, and the EKF estimation. 
		}
		\label{fig:inferq}
	\end{figure}

	In this scenario, we test our algorithm on measurements with an inaccurate noise model. An additional constant bias of 0.02 $rad/s$ is added to the simulated gyroscope measurements. The experimental result is shown in Figure.~\ref{fig:inferq}. Our proposed RLC-EKF method has shown good performance regardless of the other bias. Simultaneously, the extra unexpected noise has introduced a constant bias in the estimation of EKF. It is because such Kalman filter-based methods are derived based on the assumption that the measurement noise has a known covariance model and cannot deal with the case with an inaccurate noise model. After being compensated by an RL policy, this constraint will not exist anymore.

	\vspace{-0.5cm}
	\subsection{Results for real data\label{sec:realData}}
	
	Finally, we test the proposed algorithm on a real-world dataset. The data is collected from the Trivisio Colibri Wireless IMU with a logging rate of 100Hz. The reference measurement of the orientation is provided as ground truth from motion capture equipment by tracking the optical markers fixed to the sensor platform. The optical and IMU data have been time-synchronised and aligned beforehand. 
	
	The dataset is $100$ seconds long and split into training and inference datasets separately. The first half of the collected data is used for training and the rest for inference. We randomly selected a consecutive sequence of a length of $1000$ samples as a training episode in the training dataset. The test results are shown in Fig.~\ref{fig:my_label}. With an uncertain starting point, all the filters successfully converged to the true state and have shown similar estimation performance after convergence. Here we compare the filters' performance on the second half of the trajectory. The results can be found in Table \ref{tab:mean rmse}. 
	\begin{table}[h]
		\caption{RMSE of the orientation estimates}
		\vspace{-0.3cm}
		\label{tab:mean rmse}
		\begin{center}
			\small
			\begin{tabular}{lcccc}
				\toprule
				{RMSE} & Yaw[$rad$] & Pitch[$rad$] & Roll[$rad$]\\
				\midrule
				RLC-EKF  & 0.241& 0.020& 0.038\\
				EKF& 0.186& 0.175& 0.041 \\
				CF & 0.260& 0.019&0.041 \\
				\bottomrule
				\vspace{-0.4cm}
			\end{tabular}
			\normalsize
		\end{center}
		\vspace{-0.2cm}
	\end{table}

	\begin{figure}[h]
		\centering
		\vspace{-0.5cm}
		\includegraphics[scale=0.36]{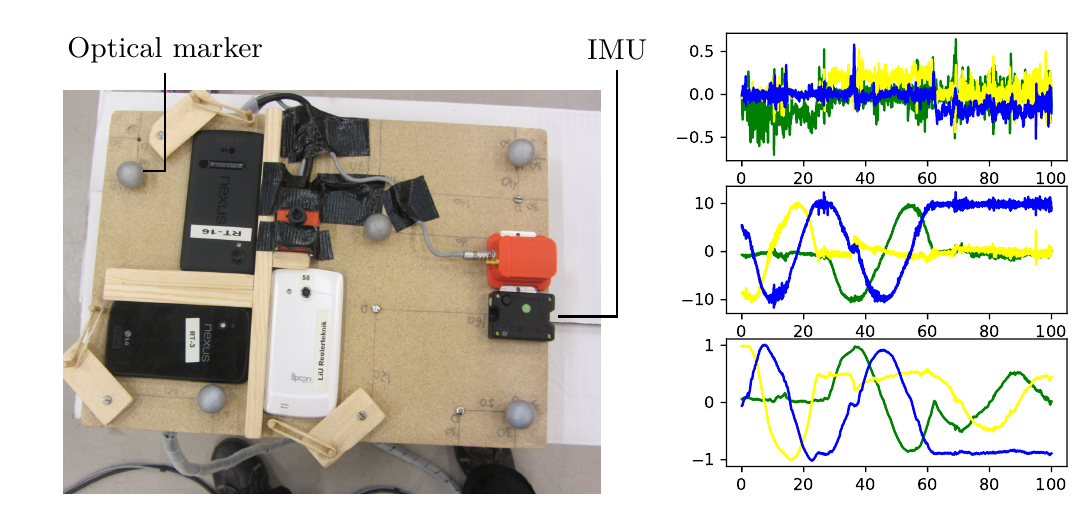}
		\caption{Real dataset (adapted from Fig.~4.2 and~4.3 in \cite{kok2017using}). Left: A snapshot of the platform for collecting real dataset  Right: Measurements from an accelerometer ($y_{\text{a},t}$, top), a gyroscope ($y_{\omega,t}$, middle) and a magnetometer ($y_{\text{m},t}$, bottom) for $100$ seconds of data collected with the IMU shown in the left figure.). 
		}
		\vspace{-0.2cm}
		\label{fig:oriEst-expSetup}
	\end{figure}

		\begin{figure}[hbt!]
		\centering
		\vspace{-0.4cm}
		\includegraphics[scale=0.38]{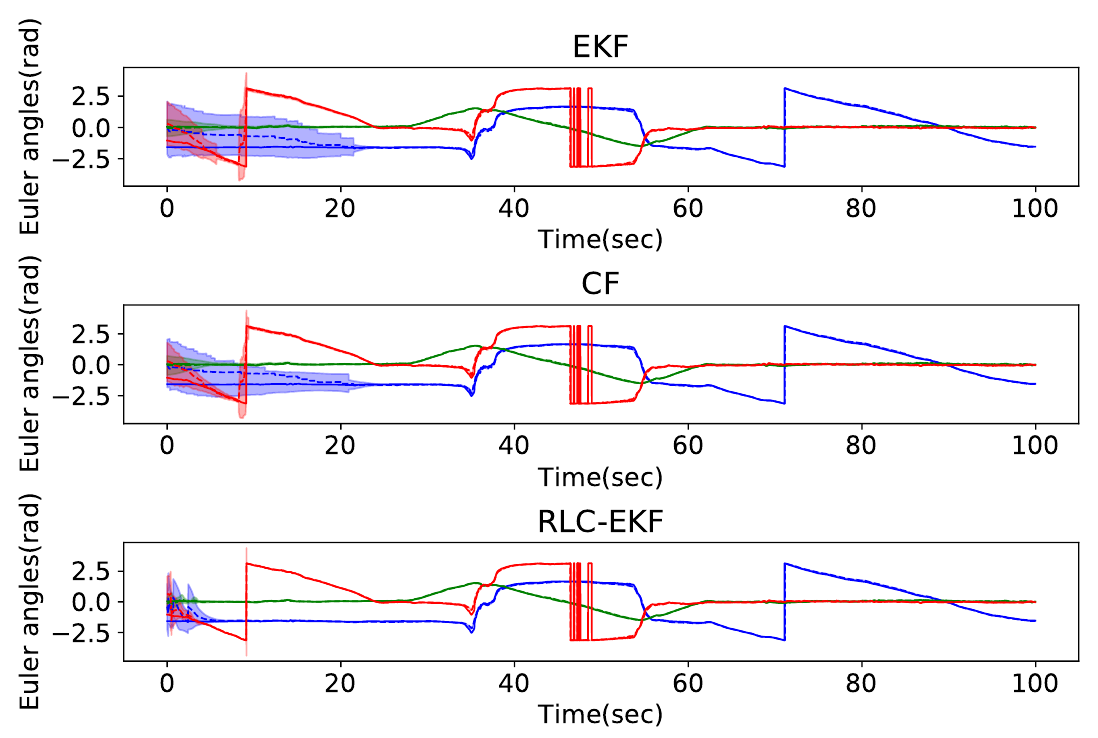}
\vspace{-0.2cm}
		\caption{Estimated Euler angles for real data. The attitude estimation is expressed in a more intuitive manner with Euler angles instead of quaternions. The solid and dashed lines correspond to the ground truth and the mean estimation respectively. The shaded areas indicate the standard deviation over 50 independent runs.}
		\label{fig:my_label}
	\end{figure}
	
 	\vspace{-0.2cm}
	\section{CONCLUSIONS}
	We have introduced a novel attitude estimation method by combining the classic EKF with a deep reinforcement learning algorithm in this work. The proposed algorithm is insensitive to (1) inaccurate initial estimate, (2) inaccurate initial gain, and (3) inaccurate noise model. The effectiveness is demonstrated on both simulated and real datasets.

	\section{ACKNOWLEDGEMENT}
	We are grateful for the help and equipment provided by the UAS Technologies Lab, Artificial Intelligence and Integrated Computer Systems Division at the Department of Computer and Information Science, Link{\"o}ping University, Sweden. We thank Gustaf Hendeby, Niklas Wahlstr{\"o}m, Hanna Nyqvist and Manon Kok who collected the real data and allow us to use. WP is supported by HUAWEI and AnKobot. YT is supported by China Scholarship Council (No. 202006890020).

	\bibliographystyle{IEEEtran}

\end{document}